\documentclass[conference]{IEEEtran}
\IEEEoverridecommandlockouts

\usepackage{cite}
\usepackage{amsmath,amssymb,amsfonts}
\usepackage{algorithmic}
\usepackage{graphicx}
\usepackage{textcomp}
\usepackage{xcolor}
\usepackage{caption}
\usepackage{subcaption}
\usepackage{url}
\usepackage{hyperref}
\definecolor{linkblue}{rgb}{0.1,0.1,0.8}
\hypersetup{
   colorlinks,
   menucolor=linkblue,
   linkcolor=linkblue,
   citecolor=linkblue,
   urlcolor=blue
}

\def\BibTeX{{\rm B\kern-.05em{\sc i\kern-.025em b}\kern-.08em
    T\kern-.1667em\lower.7ex\hbox{E}\kern-.125emX}}

\begin{document}

\newcommand{\ana}[1]{\textbf{\textcolor{blue}{Ana: #1}}} 
\newcommand{\tome}[1]{\textbf{\textcolor{green}{Tome: #1}}}
\newcommand{\carola}[1]{\textbf{\textcolor{red}{Carola: #1}}}

\title{Explainable Model-specific Algorithm Selection for Multi-Label Classification\\
\thanks{The authors acknowledge the support of the Slovenian Research Agency through research core grants No. P2-0103 and P2-0098, project grants No. J2-9230 and N2-0239, the young researcher grant No. PR-09773 to AK, as well as the EC through grant No. 952215 (TAILOR). Our work is also supported by the CNRS INS2I project RandSearch and by the ANR T-ERC project VARIATION (ANR-22-ERCS-0003).}
}

\author{\IEEEauthorblockN{1\textsuperscript{st} Ana Kostovska}
\IEEEauthorblockA{\textit{Knowledge Technologies Department,} \\
\textit{Jožef Stefan International Postgraduate School} \\
\textit{Jožef Stefan Institute}\\
Ljubljana, Slovenia \\
0000-0002-5983-7169}
\and
\IEEEauthorblockN{2\textsuperscript{nd} Carola Doerr}
\IEEEauthorblockA{\textit{CNRS, LIP6} \\
\textit{Sorbonne Universit\'e}\\
Paris, France \\
0000-0002-4981-3227}
\and
\IEEEauthorblockN{3\textsuperscript{rd} Sašo Džeroski}
\IEEEauthorblockA{\textit{Knowledge Technologies Department} \\
\textit{Jožef Stefan Institute}\\
Ljubljana, Slovenia \\
0000-0003-2363-712X}
\and
\IEEEauthorblockN{4\textsuperscript{th} Dragi Kocev}
\IEEEauthorblockA{\textit{Knowledge Technologies Department} \\
\textit{Jožef Stefan Institute}\\
Ljubljana, Slovenia \\
0000-0003-0687-0878}
\and
\IEEEauthorblockN{5\textsuperscript{th} Panče Panov}
\IEEEauthorblockA{\textit{Knowledge Technologies Department} \\
\textit{Jožef Stefan Institute}\\
Ljubljana, Slovenia \\
0000-0002-7685-9140}
\and
\IEEEauthorblockN{6\textsuperscript{th} Tome Eftimov}
\IEEEauthorblockA{\textit{Computer Systems Department} \\
\textit{Jožef Stefan Institute}\\
Ljubljana, Slovenia \\
0000-0001-7330-1902}
}

\maketitle
\thispagestyle{plain}
\pagestyle{plain}

\begin{abstract}
Multi-label classification (MLC) is an ML task of predictive modeling in which a data instance can simultaneously belong to multiple classes. MLC is increasingly gaining interest in different application domains such as text mining, computer vision, and bioinformatics. Several MLC algorithms have been proposed in the literature, resulting in a meta-optimization problem that the user needs to address: which MLC approach to select for a given dataset? 
To address this algorithm selection problem, we investigate in this work the quality of an automated approach that uses characteristics of the datasets -- so-called features -- and a trained algorithm selector to choose which algorithm to apply for a given task. For our empirical evaluation, we use a portfolio of 38 datasets. We consider eight MLC algorithms, whose quality we evaluate using six different performance metrics. We show that our automated algorithm selector outperforms any of the single MLC algorithms, and this is for all evaluated performance measures. Our selection approach is explainable, a characteristic that we exploit to investigate which meta-features have the largest influence on the decisions made by the algorithm selector. Finally, we also quantify the importance of the most significant meta-features for various domains.

\end{abstract}


\section{Introduction}
 
Multi-label classification (MLC) is a predictive modeling task that involves predicting the presence of multiple class labels which are not mutually exclusive class labels. It is present in many research areas (e.g., text categorization, where documents might be assigned multiple topics simultaneously~\cite{chen2021multi, chang2019x}, computer vision~\cite{ge2018multi, guo2019visual}, and bioinformatics~\cite{guo2019multi, zhou2020iatc}). For solving the MLC task, many different algorithms have been proposed in the literature. Previous experiments have demonstrated that there is no single algorithm that performs best on all possible datasets on a given machine learning (ML) learning task. The diversity of different MLC algorithms, therefore, poses a new meta-optimization problem: which algorithm to select when a new dataset becomes available to maximize the performance metric under consideration? This meta-optimization problem is known as the \emph{algorithm selection} (AS) problem. 

AS is a task of meta-learning~\cite{vanschoren2019meta}, where an ML model is learned to predict the best performing algorithm for new datasets (or even data instances). This requires a set of benchmark datasets that will be used for training the AS model, often referred to as a dataset portfolio; dataset characteristics presented in the same vector space for all datasets defined meta-features (this is known as meta-representation or representation of the problem landscape); and data on the performance achieved by a set of algorithms out of which the best algorithm will be selected for each dataset, often referred to as algorithm portfolio.

The AS task has already been explored for different learning tasks, based on the availability of the above-mentioned resources for the learning task: the dataset portfolio, a meta-representation for the datasets, and the performance data for an algorithm portfolio. There are several frameworks, such as OpenML~\cite{vanschoren2014openml}, ASlib~\cite{bischl2016aslib} and OPTION~\cite{kostovska2021option} that support the development of AS for different ML and optimization tasks. 

OpenML~\cite{vanschoren2014openml} is an open platform for sharing datasets, algorithms, and experimental results that can further be used for meta-learning about ML tasks. ASlib~\cite{bischl2016aslib} is a repository that contains a large number of ML datasets, together with data about the performance of the algorithms achieved on datasets especially designed and stored for performing AS. OPTION~\cite{kostovska2021option} is an ontology developed to store and make experimental data from single-objective optimization interoperable. Currently, it contains meta-features for describing single-objective optimization problem instances coming from the COCO benchmark suite~\cite{hansen2021coco}, together with performance data obtained from COCO, IOHprofiler~\cite{IOHprofiler}, and the Meta framework Nevergrad~\cite{nevergrad}. It is also worth mentioning that the DACBench library~\cite{eimer2021dacbench}, where instead of AS, data for performing algorithm configuration (AC) (i.e., selecting the best hyperparameters for a given algorithm) is stored for evolutionary computation, AI planning, and deep learning tasks. The same data can be also used in an AS learning scenario, where different hyper-parameters for the same algorithm will be treated as different algorithms.

The existence of the above-mentioned and similar libraries has led to several studies in AS~\cite{kerschke2019automated}. Tornede et al. ~\cite{tornede2022algorithm} presented a general framework for performing AS, especially focusing on meta-learning and ensemble learning methods. Shawkat and Smith ~\cite{ali2006learning} investigate AS in a classification learning scenario involving 8 different classifiers and 100 benchmark datasets. Cohen-Shapira and Rokach ~\cite{cohen2021automatic} presented an approach for AS in clustering by using supervised graph embeddings, 210 clustering datasets, and 17 clustering algorithms. Kotthoff ~\cite{kotthoff2016algorithm} provided an overview of different methods that can be used in AS for combinatorial optimization problems. Jankovic et al.~\cite{jankovic2021towards} investigated per-problem and per-instance AS for single-objective continuous optimization problems by using a meta-representation calculated from trajectory data of the algorithms and their global state variables that are changing during the optimization process. Kostovska et al.~\cite{kostovska2022per} investigated a per-run AS for single-objective continuous optimization using trajectory data of the optimization algorithms and explored the transferability of the AS results across different dataset portfolios or benchmark suites. 

However, AS for MLC is still largely unexplored. In recent work, Bogatinovski et al.~\cite{bogatinovski2022explaining} presented a study of automated algorithm performance prediction for MLC algorithms, the performance of 26 MLC algorithms was considered on a set of 40 MLC datasets described with 50 meta-features. From this data, they trained a multi-target regressor (with predictive clustering trees) to predict the performance of the algorithms with regard to several performance metrics. The main findings describe the importance of the meta-features for algorithm performance prediction with predictive clustering trees. This study motivated us to go one step beyond and instead of automated algorithm performance prediction to perform AS for MLC, i.e., select the best performing algorithm for each dataset separately. In addition, we also utilize explanation techniques to explain the AS decisions.

 The contributions of our paper can be summarized as follows:
\begin{itemize}
    \item Using ML, we trained MLC algorithm selector for six different evaluation measures on a portfolio comprised of eight different MLC algorithms. The results show that the algorithm selector provides better results across the different evaluation metrics when compared to the best single algorithms, confirming our hypothesis that state-of-the-art algorithm selection techniques provide a promising alternative to a manual choice. 
    \item To provide explanations for the choices made by the algorithm selector, we investigate the impact of the MLC meta-features on the automatic algorithm performance prediction for each algorithm separately and combine the contribution of the MLC meta-features based on the selected best algorithms. 
    \item We present domain-specific explanations for the algorithm selector to explore the differences in the landscape of MLC meta-features when training an algorithm selector for each domain separately. 
\end{itemize}

\textbf{Outline of the Paper:} In Section~\ref{sec:background}, we introduce the problem of algorithm selection and provide a background of the MLC meta-features used to characterize the landscape of MLC datasets. Section~\ref{sec:setup} provides details of the experimental setup and the quality estimation of the MLC selector. In Section~\ref{sec:results}, we discuss the results of our empirical evaluation. The domain-specific explanations of the algorithm selector are given in Section~\ref{sec:explainable}. Finally, Section~\ref{sec:conclusions} concludes our study by highlighting the main contributions and discussing possible directions for future work. 

\textbf{Availability of data and code for reproducibility:} Source code, performance and landscape data, results, and figures produced for our study are available in the Zenodo repository accompanying this paper~\cite{kostovska_ana_2022_6829671}. Please note that the repository also contains additional information, not described here for reasons of space limitation.

\section{Background}
\label{sec:background}
\subsection{Algorithm Selection}
Algorithm selection is a meta-algorithmic design technique that addresses the problem of choosing a well-performing algorithm from a finite, performance-complementary algorithm portfolio, on a per-instance basis. There are different strategies for building an algorithm selector (e.g., parallel algorithm portfolios, algorithm schedules, or ML-based automated algorithm selection). 
In ML, algorithm selection can be treated as a classification or regression task. When treated as multi-class classification, the input is the landscape features (or meta-features) of the dataset instance, and the output (or target) is the best-performing algorithm out of an algorithm portfolio. Alternatively, pairwise classification can be employed, where the relative performance of pairs of algorithms is compared and the algorithm with the most ``wins'' is selected~\cite{rijn2015fast}. When treated as a regression task, separate regression models for performance prediction are trained for each algorithm in the portfolio. The algorithm with the best-predicted performance is selected. 

To estimate the quality (or the performance gain) of an algorithm selector, two baselines are commonly used in the literature~\cite{kerschke2019automated}: (i) the performance of the algorithm that maximizes mean performance on the dataset portfolio (called single best solver (SBS)); (ii) the oracle performance or the virtual best solver (VBS) -- a hypothetical, perfect selector that chooses the best performing algorithm for each dataset instance.

\subsection{MLC Meta-features}
Meta-learning is a sub-field of ML concerned with learning from past experiences i.e., data on past machine learning experiments, commonly referred to as meta-data~\cite{brazdil2008metalearning}. The main goal of meta-learning is to enable the automation of parts of the machine learning pipeline, i.e., the selection of machine learning algorithms that are most suitable for a given dataset. The meta-data usually includes dataset characteristics (or meta-features) that are relevant to the learning task. These meta-features allow for grouping the datasets according to their similar characteristics, which can be used for transferring knowledge from one dataset to other datasets in the same group. 

Defining a proper set of meta-features for specific learning tasks has been a question of interest for data scientists. Moyano et al. ~\cite{Moyano2017} defined a list of meta-features specific for multi-label classification datasets, categorized into five meta-feature groups: (1) dimensionality, e.g., number of features, number of labels, number of instances; (2) label distribution, e.g., frequency, cardinality and density of labels; (3) label imbalance, e.g., mean of inter-class imbalance ratio; (4) labels relationship, e.g., proportion of distinct labelsets; and (5) attribute metrics, e.g., number of binary attributes. In this study, we use these meta-features to represent the landscape of MLC datasets. 

\section{Experimental Setup}
\label{sec:setup}
This section describes the experimental setup, which includes the description of the dataset portfolio, the landscape data associated with the datasets, the MLC algorithm portfolio, and the performance data. Following that, we present details on how the regression models that are the basis for building the algorithm selector are trained. Finally, we describe how we build the algorithm selector. 

\subsection{Dataset Portfolio and Landscape Data} 
The dataset portfolio consists of 38 MLC datasets that have previously been used in various studies for benchmarking MLC methods. The datasets come from five different application domains (i.e., text, multimedia, bioinformatics, medical, and chemistry). Further, the portfolio covers datasets with a diverse number of labels (4-274), data instances (139-17190), and descriptive features (33-49060). Our ML pipeline for building the algorithm selector relies on having meta-descriptors (or meta-features) of the MLC datasets in order to train the regression models. For that purpose, we reuse a set of 63 MLC meta-features that have already been proposed in the literature~\cite{moyano2017mlda} and are shown to provide promising results when predicting algorithm performance~\cite{bogatinovski2022explaining}. We reduce the initial set of 63 meta-features to 17 by calculating Pearson correlation pairwise and removing one of the features in the pair with a correlation larger than 0.75. All datasets and the related MLC meta-features have been downloaded from the publicly available MLC data catalogue~\cite{kostovska2022catalogue}.\\

\subsection{MLC Algorithm Portfolio and Performance Data} 
The performance data we use here comes from a comprehensive comparative study of MLC algorithms~\cite{bogatinovski2022comprehensive}. The study evaluates 26 MLC algorithms over 42 datasets (including the 38  datasets mentioned above) using 18 predictive performance evaluation measures and two efficiency performance measures.

In this study, we are only concerned with selecting the algorithm that performs best and we ignore the efficiency component, i.e., we ignore the two efficiency performance measures. Since the evaluation measures are correlated, we remove the evaluation measures with a Pearson correlation larger than 0.90. This leaves us with six evaluation measures: average precision, macro F1, one error, AUROC, Hamming loss, and micro precision.

Next, to create a portfolio of MLC algorithms with complementarity in their performance, for each combination of dataset and evaluation measure, we count the number of times a given algorithm is performing the best. In the portfolio, we include the algorithms that performed the best on at least eight out of the 38 datasets for any of the six evaluation metrics. Following this criteria, the final MLC method portfolio consists of eight algorithms (i.e., AdaBoost ~\cite{Schapire2000}, Calibrated Label Ranking (CLR)~\cite{Brinker2006, Frnkranz2008}, Deep Belief Networks (DEEP4 version)\cite{Salakutinov2006}, Hierarchy of Multi-label Classifiers (HOMER)~\cite{Tsoumakas2008}, Multi-label Adaptive Resonance Associative Map (MLARM)~\cite{Sapozhnikova2009}, The method of Pruned Sets (PSt)~\cite{Read2008},  Binary relevance with random forest (RFDTBR)\cite{tsoumakas2007multi}, and Random Forest of Predictive Clustering Trees (RFPCT)~\cite{Kocev20https://obofoundry.org/principles/fp-000-summary.html11,Madjarov2012}).

\subsection{Regression Models}

\begin{table}[t]
\begin{center}
\caption{RF hyperparameter names and their corresponding values considered in the grid search.}
\label{tab:hyperparameters}
\begin{tabular}{ cc }
\hline
 Hyperparameter & Search space \\ 
\hline
 n\_estimators & $[50, 100]$\\
 max\_features & $[\textsc{auto}, \textsc{sqrt}, \textsc{log2}]$ \\ 
 max\_depth & $[4,8,15, \textsc{None}]$ \\
 min\_samples\_split & $[2, 5, 10]$ \\
 \hline
\end{tabular}
\end{center}
\end{table}

For training the regression models we considered two scenarios: 
\begin{itemize}
    \item Single target regression (STR) -- we train a separate regression model for each evaluation metric per MLC algorithm. That leaves us with 48 different regression models (8 MLC methods $\times$ 6 evaluation metrics).
    \item Multi-target regression (MTR) -- we train one regression model per MLC method where we simultaneously try to predict the performance of the method according to the six evaluation metrics. In this scenario, we train eight different regression models since we have an MLC method portfolio of size eight. 
\end{itemize}

The regression models are built with the Random Forest (RF) algorithm as implemented in the Python package \texttt{scikit-learn}~\cite{pedregosa2011scikit}. The RF hyperparameters are tuned using the grid search methodology. We tune four different RF hyperparameters: (1) \emph{n\_estimators} -- the number of trees in the random forest; (2) \emph{max\_depth} -- the maximum depth of the trees ; (3) \emph{max\_features} -- the number of features used for making the best split ; and (4) \emph{min\_samples\_split} -- the minimum number of samples required for splitting an internal node. Thus, for each trained model, we consider 72 different model candidates. The full list of tuned hyperparameters and the corresponding search spaces are given in Table~\ref{tab:hyperparameters}. 

\begin{figure}
\centerline{ \includegraphics[width=\linewidth]{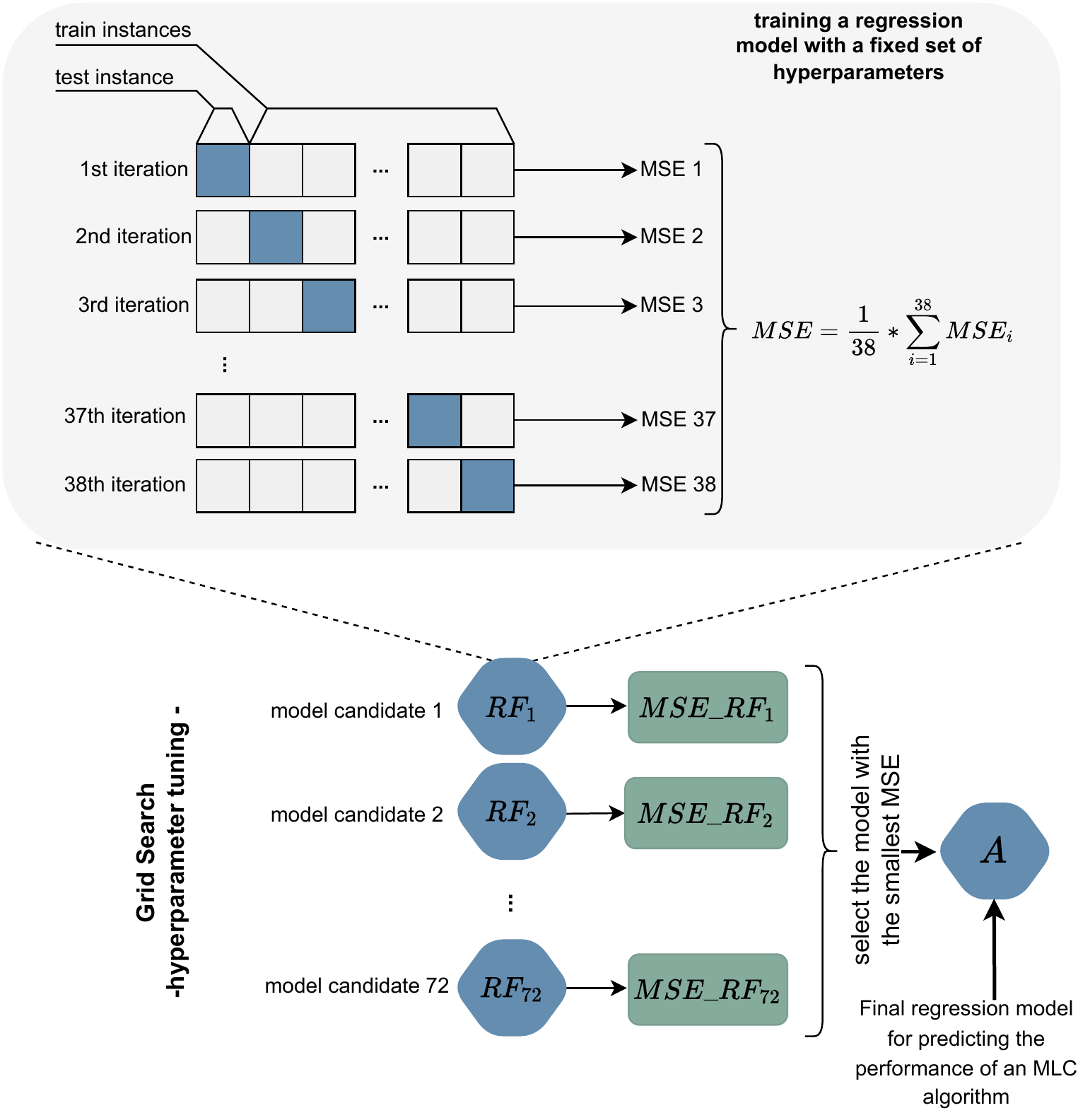}}
\caption{An illustration of the process of training a regression model for predicting the performance for an MLC algorithm.}
\label{fig:grid}
\end{figure}

The regression models are evaluated with the \emph{leave-one-instance-out} strategy, where an instance is one MLC dataset. Since we have 38 MLC datasets, we perform training 38 times where we hold one instance for testing and 37 for training. Finally, we compute the mean squared error and average it across all test instances (see Figure~\ref{fig:grid}.  

\begin{figure}
\centerline{ \includegraphics[width=0.9\linewidth]{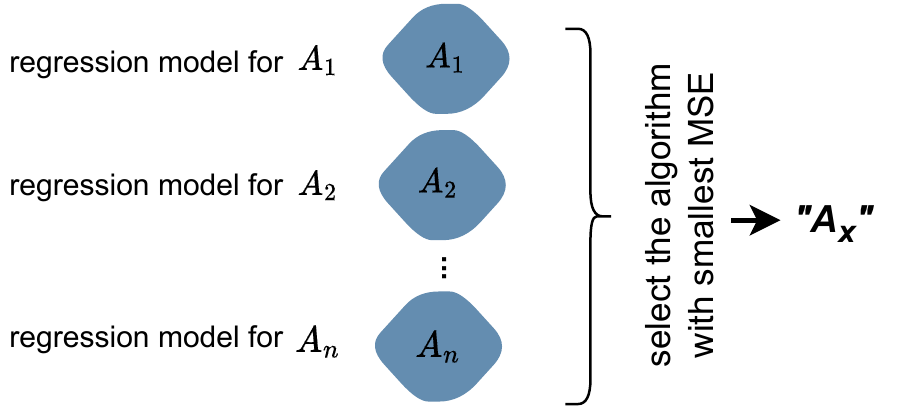}}
\caption{An illustration of the process of selecting the best performing algorithm for an algorithm portfolio of size $N$.}
\label{fig:algo}
\end{figure}

\subsection{Construction and Quality Estimation of the MLC Algorithm Selector}
After training the regression models for performance prediction of each of the MLC algorithms separately, we select the best performing algorithm (the one with the best-predicted performance) on every dataset, w.r.t. each evaluation metric (see Figure~\ref{fig:algo}). Note that the selection process is the same for the single and multi-target settings and the two settings only differ in the prediction step. 

\begin{figure*}
\centerline{ \includegraphics[width=0.7\linewidth]{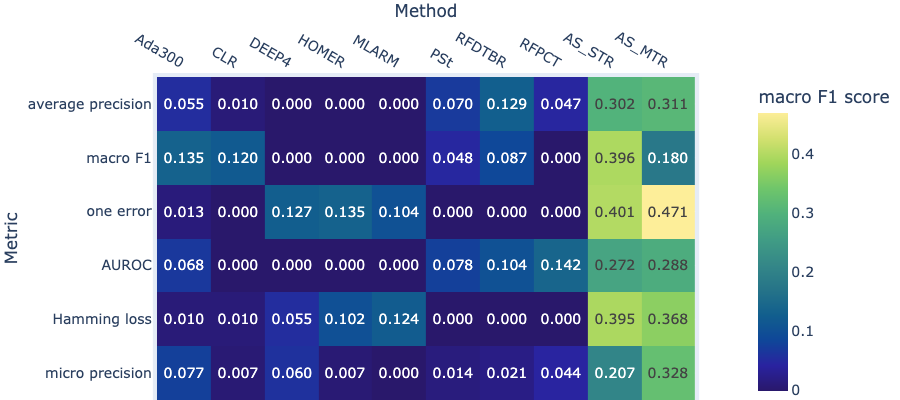}}
\caption{A heatmap showing the f1-macro of the multi-class classification for each of the methods chosen as single best solvers and for the two algorithm selectors (single and multi-target) across the six different MLC evaluation metrics.}
\label{fig:f1-macro}
\end{figure*}

To estimate the quality (or the performance gain) of the algorithm selector, we take two approaches: (i) we treat the algorithm selection as a multi-class classification problem and report the macro f1 score, and (ii)  we compute the absolute difference between the target performance value reached by the selected best algorithm and the true best algorithm.

\begin{figure*}
     \centering
     \begin{subfigure}[b]{0.48\textwidth}
         \centering
         \includegraphics[width=\textwidth]{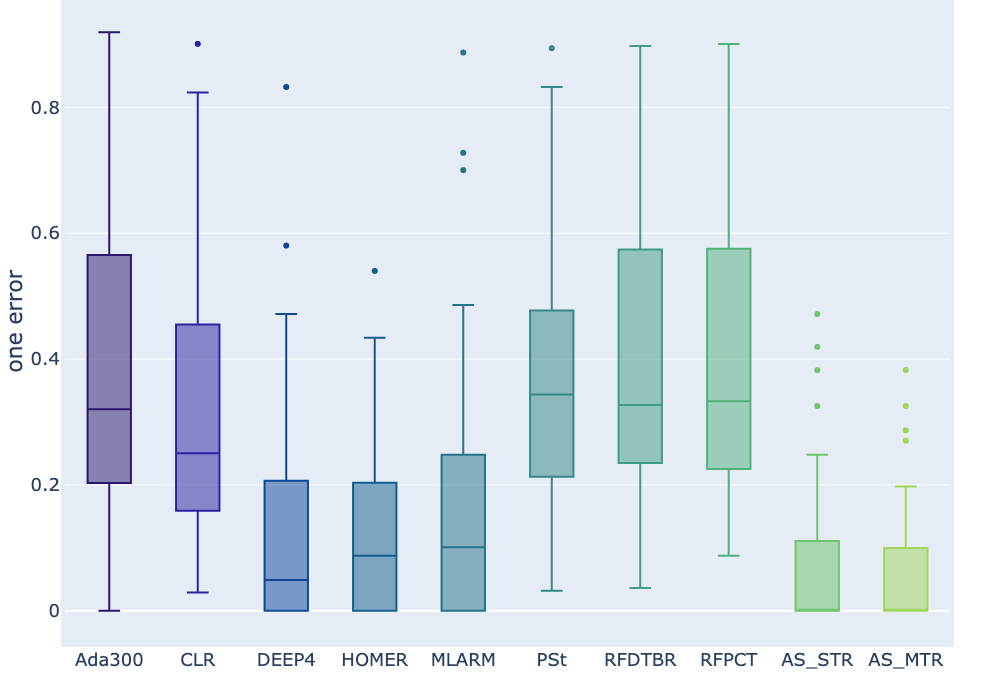}
     \end{subfigure}
     \hfill
     \begin{subfigure}[b]{0.48\textwidth}
         \centering
         \includegraphics[width=\textwidth]{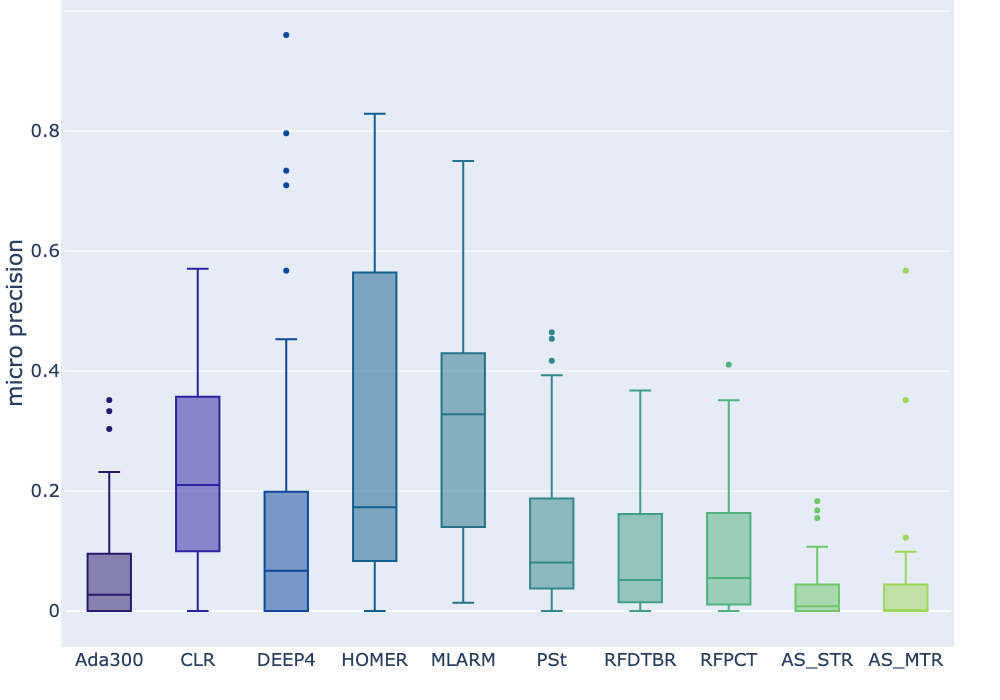}
     \end{subfigure}
     \hfill
     \begin{subfigure}[b]{0.48\textwidth}
         \centering
         \includegraphics[width=\textwidth]{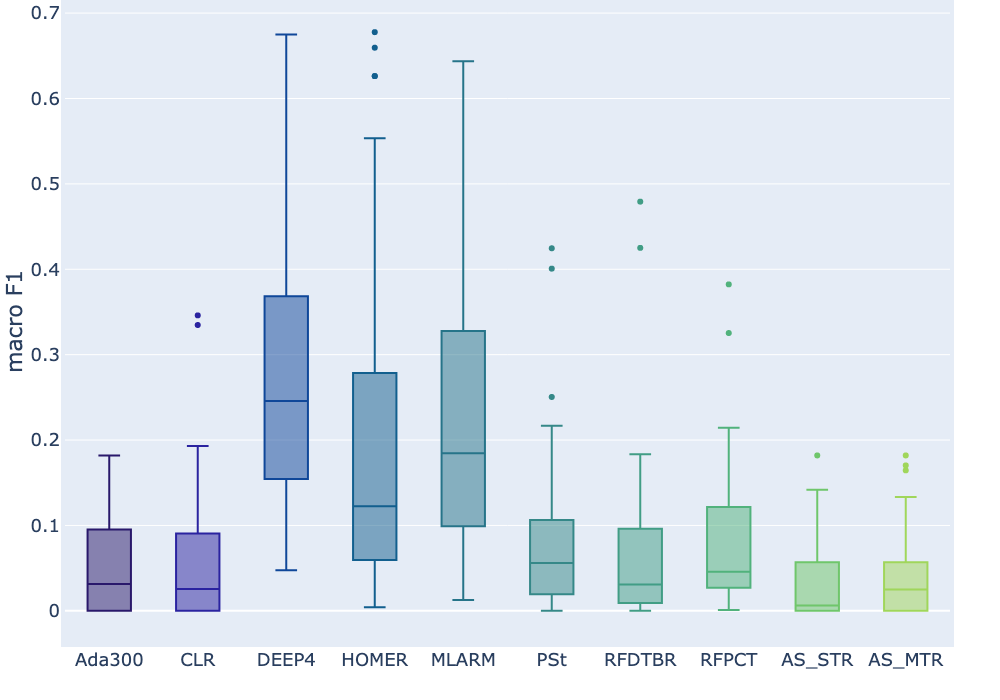}
     \end{subfigure}
     \hfill
     \begin{subfigure}[b]{0.48\textwidth}
         \centering
         \includegraphics[width=\textwidth]{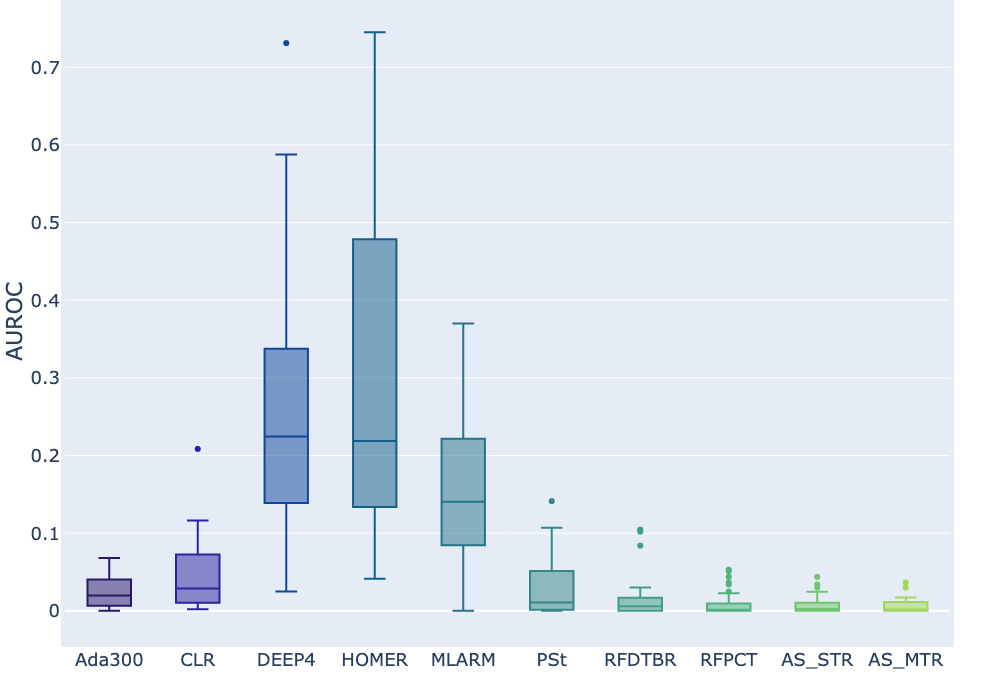}
     \end{subfigure}
     \hfill
     \begin{subfigure}[b]{0.48\textwidth}
         \centering
         \includegraphics[width=\textwidth]{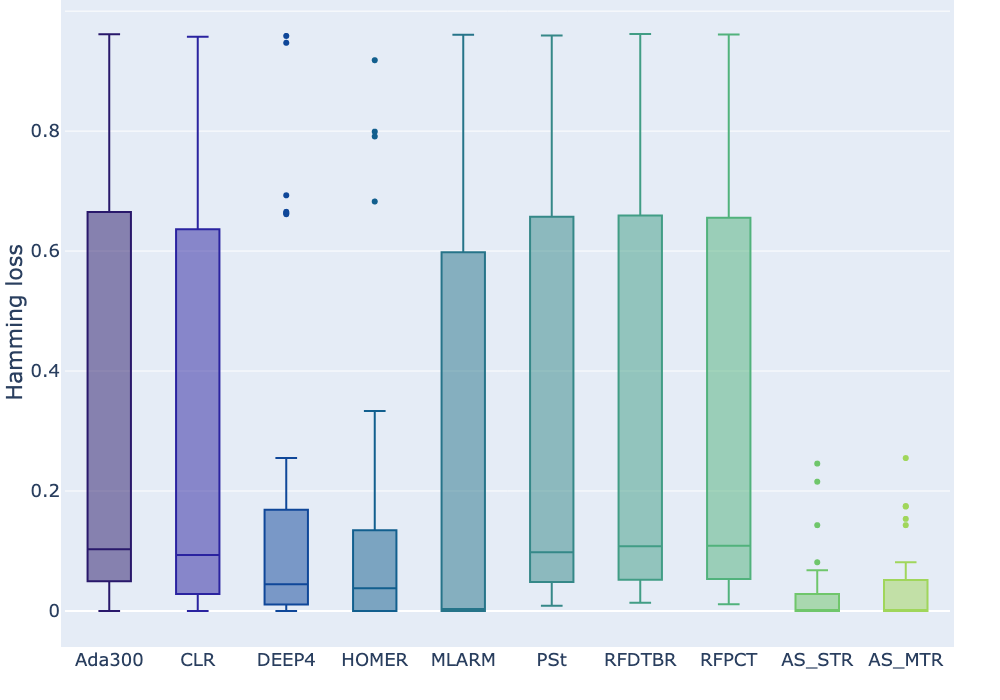}
     \end{subfigure}
     \hfill
     \begin{subfigure}[b]{0.48\textwidth}
         \centering
         \includegraphics[width=\textwidth]{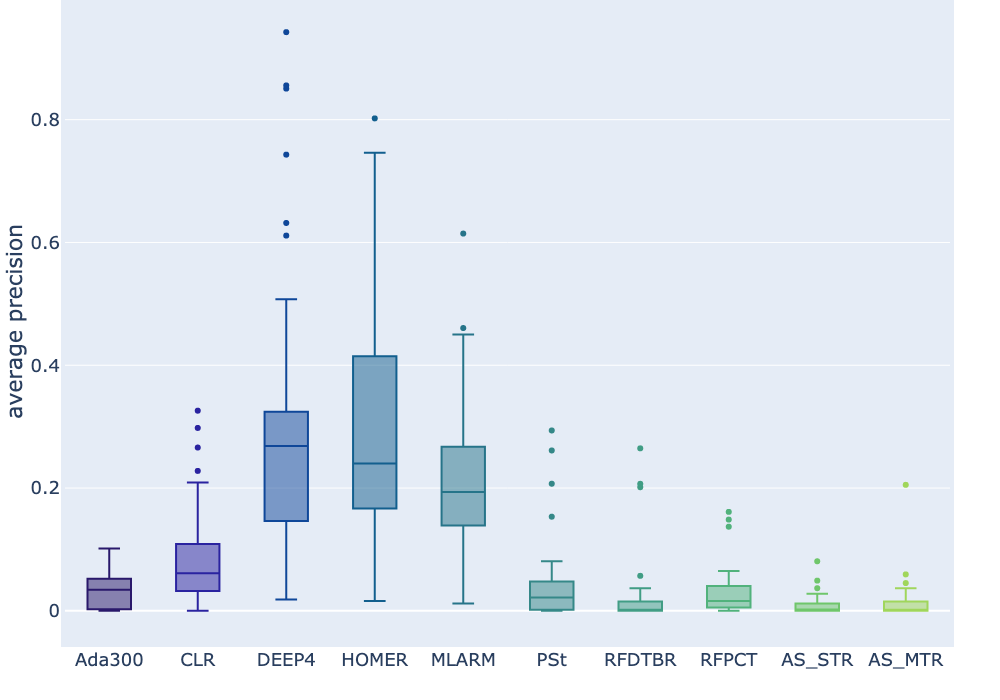}
     \end{subfigure}
    \caption{Boxplots depicting the absolute difference in the achieved target value by the single best solver (SBS) and the target value of the virtual best solver (VBS) for each of the evaluation metrics.}
    \label{fig:boxplots}
\end{figure*}

\section{Evaluation Results}
\label{sec:results}

We first present the results of the algorithm selector where we estimate its performance on the corresponding multi-class classification problem. Fig.~\ref{fig:f1-macro} provides the heatmap of the macro F1 scores with each of the MLC algorithms in the portfolio as chosen/predefined as the single best solver and when we take predictions of the two algorithm selectors we build (single and multi-target) for each evaluation metric. 

First of all, we should note that the results are highly dependent on the evaluation metric. For example, RFPCT can be considered the single best solver according to the AUROC metric (it has a 0.142 macro F1 score -- the highest when compared to the other algorithms in the portfolio). However, for 3 out of the 6 evaluation metrics, it has a macro F1 score of 0, which means that in the available performance data, this algorithm did not perform the best for any of the datasets (for the 3 metrics considered). 

Further, the results indicate that there is a performance gain when using our methodology for algorithm selection. For instance, the macro F1 score increases from 0.142 for the single best solver to 0.272/0.288 for the algorithm selector build in the single/multi-target setting, when considering AUROC as an evaluation metric. The single/multi-target algorithm selector achieves the best performance, 0.401/0.471, for the ONE ERROR evaluation metric. 

Since the evaluation measures (the multiple targets in our experimental setup) are correlated, we wanted to investigate whether the performance of the selector improves when we train multi-target regression models and make predictions simultaneously for the multiple targets (or evaluation measures). We can observe that for 4 out of the 6 evaluation measures considered in the study, the performance indeed improves. However, the performance slightly drops when considering Hamming loss as the evaluation metric (from 0.395 to 0.368) and there is a significant drop in performance in the case of macro F1 (from 0.396 to 0.108). This could be explained by the fact that Hamming loss has a lower correlation with the other evaluation metrics. However, even though macro F1 has strong a correlation with the other metrics, performance suffers in the multi-target setting. We intend to look into this phenomenon in more detail in subsequent research. All correlation plots can be found at ~\cite{kostovska_ana_2022_6829671}.

\begin{figure*}
\centerline{ \includegraphics[width=0.7\linewidth]{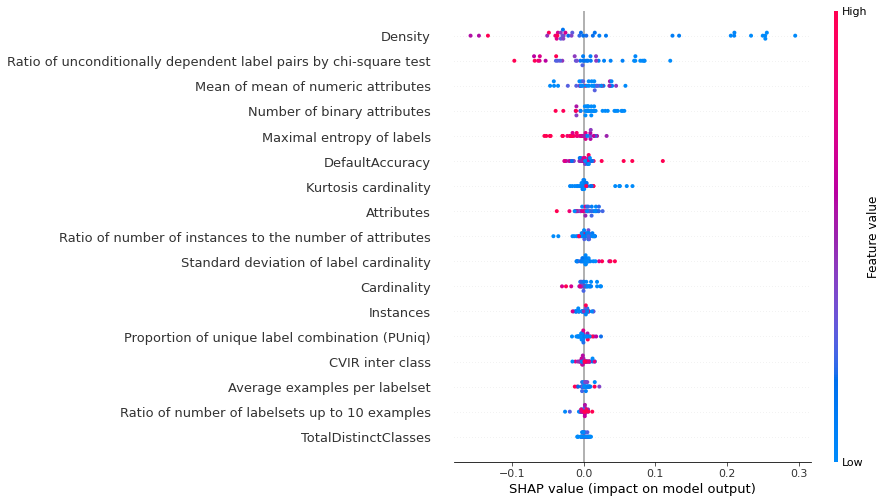}}
\caption{Summary plot of the Shapley values obtained for the single-target algorithm selector and the \textit{one error} evaluation metric across all datasets. }
\label{fig:summary}
\end{figure*}

Treating the algorithm selection as a multi-class classification problem might not give us an accurate estimate of the performance gain. It can happen that the selected algorithm is not the same as the virtual best solver, but has very similar performance. The opposite is also true, the performance difference between the selected and the virtual best solver can be very large. To take into consideration this discrepancy in the target value, we also provide boxplots that depict the absolute difference between the target value reached by the selected best and the virtual best solver (see Fig.~\ref{fig:boxplots}). Here, we can also observe that the algorithm selector improves performance when comparing it to the single best solvers. However, the performance gain (the gap closed between the VBS and the SBS) largely depends on the evaluation metric. More specifically, for the one error, micro precision, macro F1, and Hamming loss we have a large performance gain, whereas for the AUROC and average precision the performance of the algorithm selector is very similar to the one of RFDTBR and RFPCT taken as single best solvers.

\section{Explainable Algorithm Selection}
\label{sec:explainable}

\begin{figure*}
     \centering
     \begin{subfigure}[b]{0.48\textwidth}
         \centering
         \includegraphics[width=\textwidth]{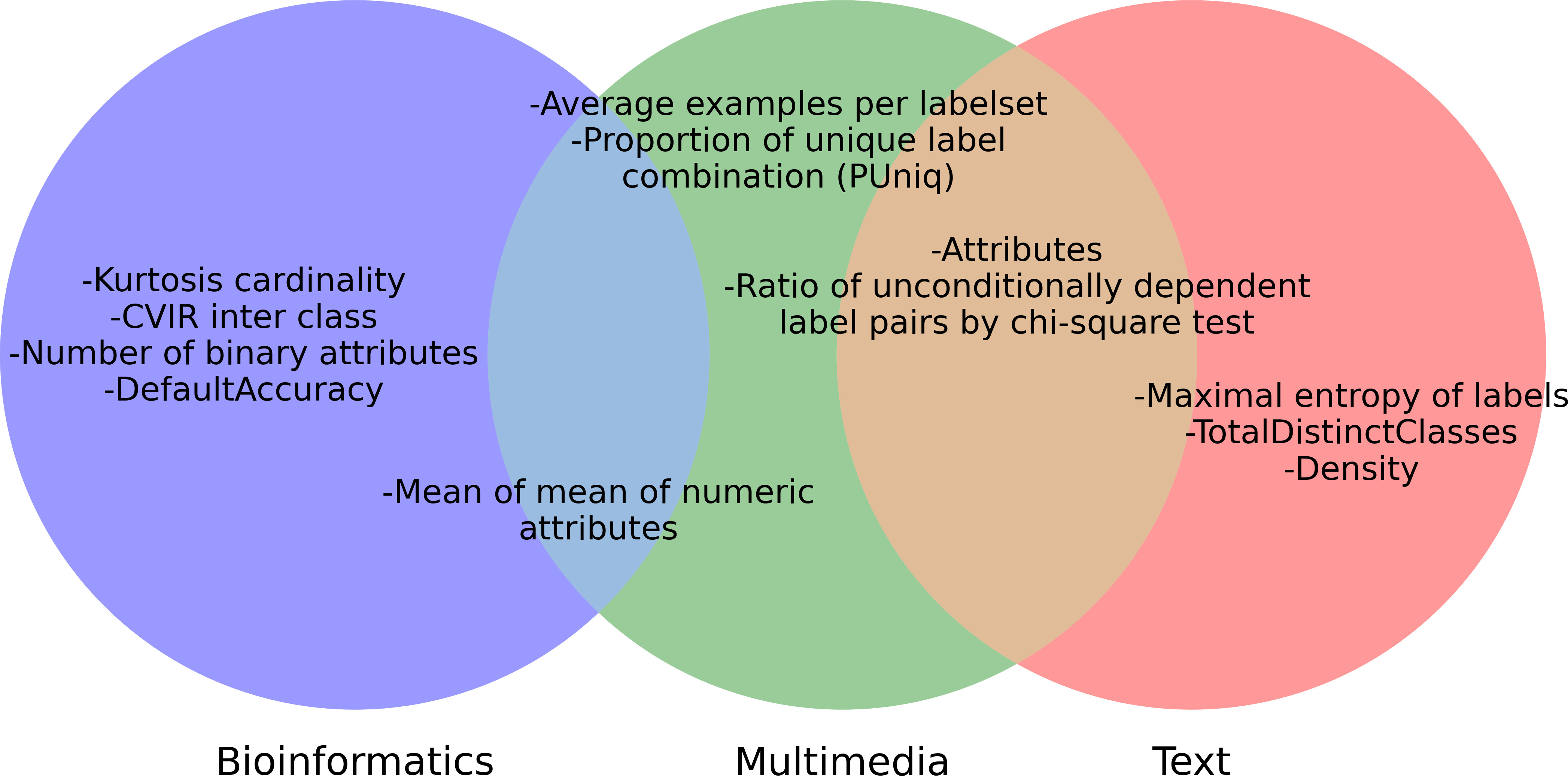}
         \subcaption{one error}
     \end{subfigure}
     \hfill
     \begin{subfigure}[b]{0.48\textwidth}
         \centering
         \includegraphics[width=\textwidth]{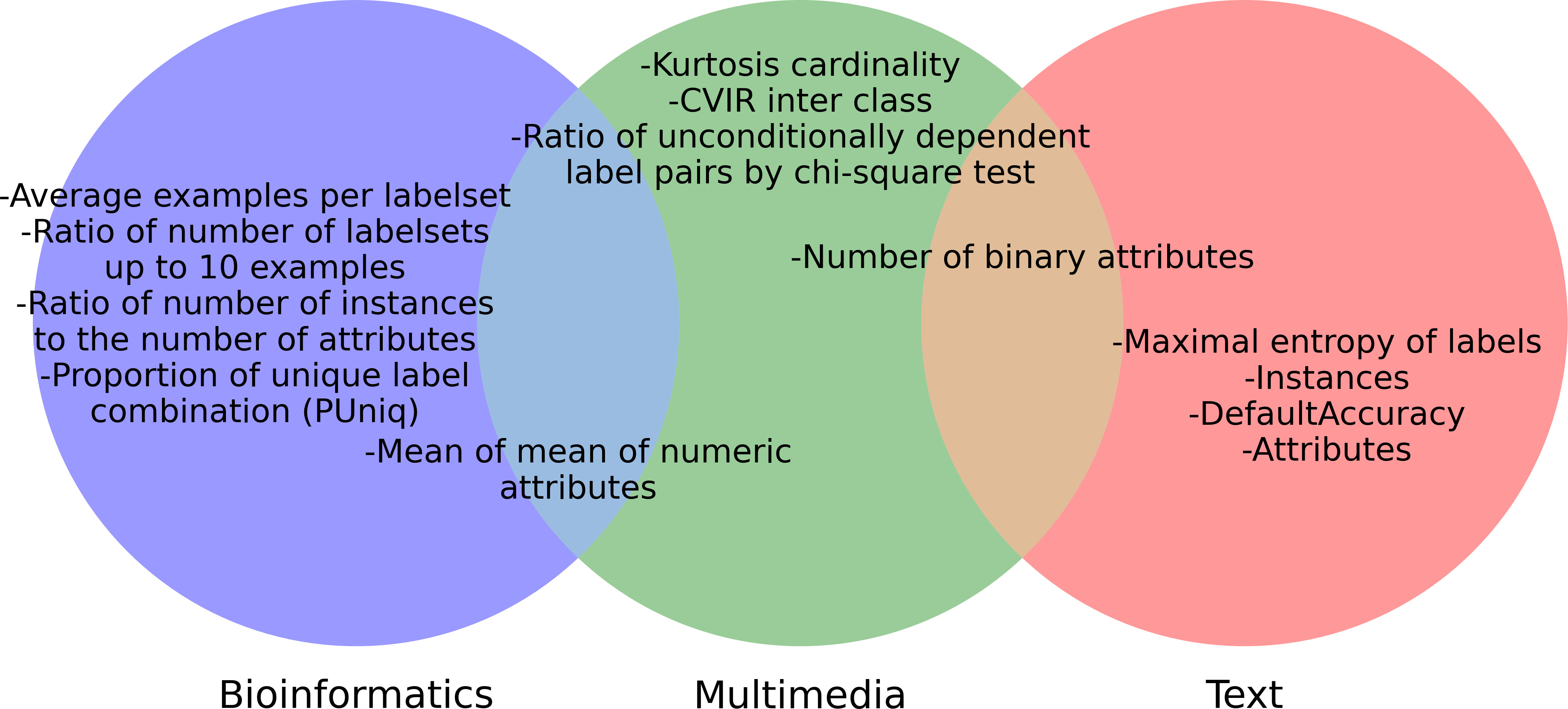}
         \subcaption{AUROC}
     \end{subfigure}
     \hfill
    \caption{Venn diagrams with the top 5 most important MLC meta-features for the \textit{Bioinformatics}, \textit{Multimedia}, and \textit{Text} domains w.r.t. the (a) \textit{one error} and (b) \textit{AUROC} evaluation measures.}
    \label{fig:venn}
\end{figure*}
To provide better explanations of the algorithm selection models, we proceed with the calculation of the Shapley values, i.e. the MLC meta-feature importance scores for the AS models. The Shapely values quantify the marginal contribution of the input features on the predictions made by the predictive AS model~\cite{molnar2020interpretable}. However, the calculation of the Shapley values is computationally expensive as it considers each possible combination of features (a power-set of features) to determine their contribution to the prediction process. Therefore, we apply the SHAP (SHapley Additive exPlanations) algorithm~\cite{lundberg2017unified}. 

\subsection{General explanations}
In order to obtain the Shapley values for the algorithm selector, for each MLC dataset, we first check which is the selected algorithm and then get the Shapley values by using the regression model trained for the selected algorithm. Next, we provide a summary plot that illustrates the positive and negative relationships of the meta-features with the quality of the prediction. Each dot in the plots represents a dataset and the MLC meta-features are listed in descending order of importance. The colors used indicate the magnitude of the MLC meta-feature value (red representing higher values and blue representing lower values). Finally, the position on the horizontal axis presents the impact of the MLC meta-feature value on the prediction of the target. Fig.~\ref{fig:summary} shows the summary plot for the single target algorithm selector with respect to the \textit{one error} evaluation metric. We can see that \textit{Density}, \textit{Ratio of unconditionally dependent label pairs by chi-square test} and \textit{Mean of mean of numeric attributes}, appear as the top 3 most important meta-features. Because of the limited number of pages available, the summary plots for the other evaluation metrics and for the multi-target algorithm selector are not included here but are available at our Zenodo repository~\cite{kostovska_ana_2022_6829671}.

\subsection{Domain-specific Explanations}
Next, we investigate the MLC meta-features' importance in the predictive performance for the algorithm selector at the level of domains from which datasets originate. The 38 MLC datasets taken into account in this study come from 5 distinct application domains, including text (15), multimedia (5), bioinformatics (15), medical (2), and chemistry (1)). In the following analysis, for simplicity, we focus only on the text, multimedia, and bioinformatics domains as they appear as the most frequent domains in our dataset portfolio. To obtain domain-specific explanations, we take the Shapley values of the MLC meta-features, group them by domain and compare the feature importance ranks to see if they vary across the different domains.

Figure~\ref{fig:venn} depicts Venn diagrams with the top 5 most important MLC meta-features for the text, multimedia and bioinformatics domain w.r.t. the \textit{one error} and \textit{AUROC} evaluation measures. An interesting observation is that the most important MLC meta-features do not overlap for the bioinformatics and the text domain. The same pattern can be observed for the other evaluation measures as well (check our Zenodo repository~\cite{kostovska_ana_2022_6829671} that includes all resources and generated figures). This might be due to the nature of the datasets, but it is something we that should be investigated further.

\section{Conclusions}
\label{sec:conclusions}
In this paper, we have investigated the potential of automated algorithm selection for the multi-label classification (MLC) learning task. We have trained random forest models in both single- and multi-target scenarios, to predict the performance of the algorithms with regard to six performance metrics. We used 38 datasets, represented with 17 selected meta-features and an algorithm portfolio of eight MLC algorithms. The results of the performance prediction were used to select the best algorithm for each dataset separately. The evaluation results showed that algorithm selection yields performance gains over the scenario, which uses a single MLC algorithm (the one which is, on average, the best for all datasets) regardless of the evaluation measure that is predicted. By combining the explanations obtained for each performance prediction model separately, we have additionally provided explanations about which meta-features most influence the decisions made by the algorithm selector (AS).

We come to the conclusion that there is an overlap between the meta-features from the multimedia datasets and the datasets from the bioinformatics and text domains separately after looking at the explanations provided for various datasets and examining them based on the dataset domain. The most significant meta-features, however, that are used to help the algorithm choose between datasets from bioinformatics and text, do not overlap.

For our future work, we intend to build on this work by incorporating additional datasets for the MLC learning task. We will also look at how the datasets, described by their meta-feature representation are distributed in the landscape and choose datasets that are evenly distributed in the landscape space that will be used to train the AS. By doing this, we hope to reduce the bias of the AS toward certain types of dataset distribution landscapes. Finally, we consider presenting a more thorough analysis of the justifications offered by the AS and their intersection with various dataset domains and evaluation metrics.

\section*{Acknowledgment}
The authors acknowledge the support of the Slovenian Research Agency through research core grants No. P2-0103 and P2-0098, project grant No. N2-0239, and the young researcher grant No. PR-09773 to AK, as well as the EC through grant No. 952215 (TAILOR). Our work is also supported by Paris Ile-de-France region, via the DIM RFSI AlgoSelect project and via a \href{http://species-society.org/scholarships-2022/}{SPECIES scholarship} for Ana Kostovska.

\bibliographystyle{IEEEtran}
\bibliography{references}
\end{document}